\documentclass[letterpaper, 10 pt, conference]{ieeeconf}

\IEEEoverridecommandlockouts
\usepackage{cite}
\usepackage{url}
\usepackage{amsmath,amssymb,amsfonts}
\usepackage{graphicx}
\usepackage{textcomp}
\usepackage{microtype}
\usepackage{xcolor}

\usepackage{booktabs,array,tabularx}
\usepackage{makecell}
\usepackage{cuted}
\usepackage{caption}
\usepackage{tikz}
\usepackage[hidelinks]{hyperref}

\title{\LARGE\bf SAKI: Skill Assembly and Kinematic Imitation from Human Videos for Long-Horizon Mobile Manipulation}
\author{Yijie Lu$^{1,\dagger}$, James Zhao$^{1,\dagger}$, and Weiming Zhi$^{1,2,*}$\\[2pt]
\small $^{1}$School of Computer Science and $^{2}$Australian Centre for Robotics, The University of Sydney, Australia\\
\small $^{\dagger}$Equal contribution. $^{*}$Corresponding author: \href{mailto:Weiming.Zhi@sydney.edu.au}{\texttt{Weiming.Zhi@sydney.edu.au}}.\\
\small Project website: \url{https://aus.bot/research/saki/}}
\hypersetup{
  pdftitle={SAKI: Skill Assembly and Kinematic Imitation from Human Videos for Long-Horizon Mobile Manipulation},
  pdfauthor={Yijie Lu, James Zhao, Weiming Zhi},
  pdfsubject={Robotics; long-horizon mobile manipulation},
  pdfkeywords={human video imitation, skill assembly, mobile manipulation, whole-body planning}
}
\begin{document}
\maketitle
\thispagestyle{empty}
\pagestyle{empty}

\suppressfloats[t]
\begin{abstract}
Learning from human videos offers a promising route to acquiring diverse manipulation skills. Extending this capability beyond tabletop settings to long-horizon mobile manipulation requires adapting and composing demonstrated interactions across changing scenes and robot configurations. We present \emph{Skill Assembly and Kinematic Imitation} (SAKI), a framework connecting human-video skill acquisition, cross-demonstration assembly and closed-loop whole-body execution. SAKI prepares reusable object-centric skills that preserve task-critical interactions while allowing transfer paths to adapt. Given a goal and supplied task dependencies, it selects and orders skills, binds their object roles to the current scene, and carries scene estimates and robot configuration between successive skills. Whole-body kinematic imitation generates coordinated base, arm and gripper motion. During execution, persistent object estimates maintain task references across viewpoint changes, while visual feedback updates remaining trajectories. Real-robot experiments demonstrate skill reuse across layouts and the composition of independently demonstrated interactions into continuous mobile tasks, including tidying and wiping. Ablation results show that task-conditioned reference preparation substantially improves long-horizon task completion with whole-body optimisation and visual feedback held fixed.
\end{abstract}

\begin{figure}[t]
\centering
\includegraphics[width=\linewidth]{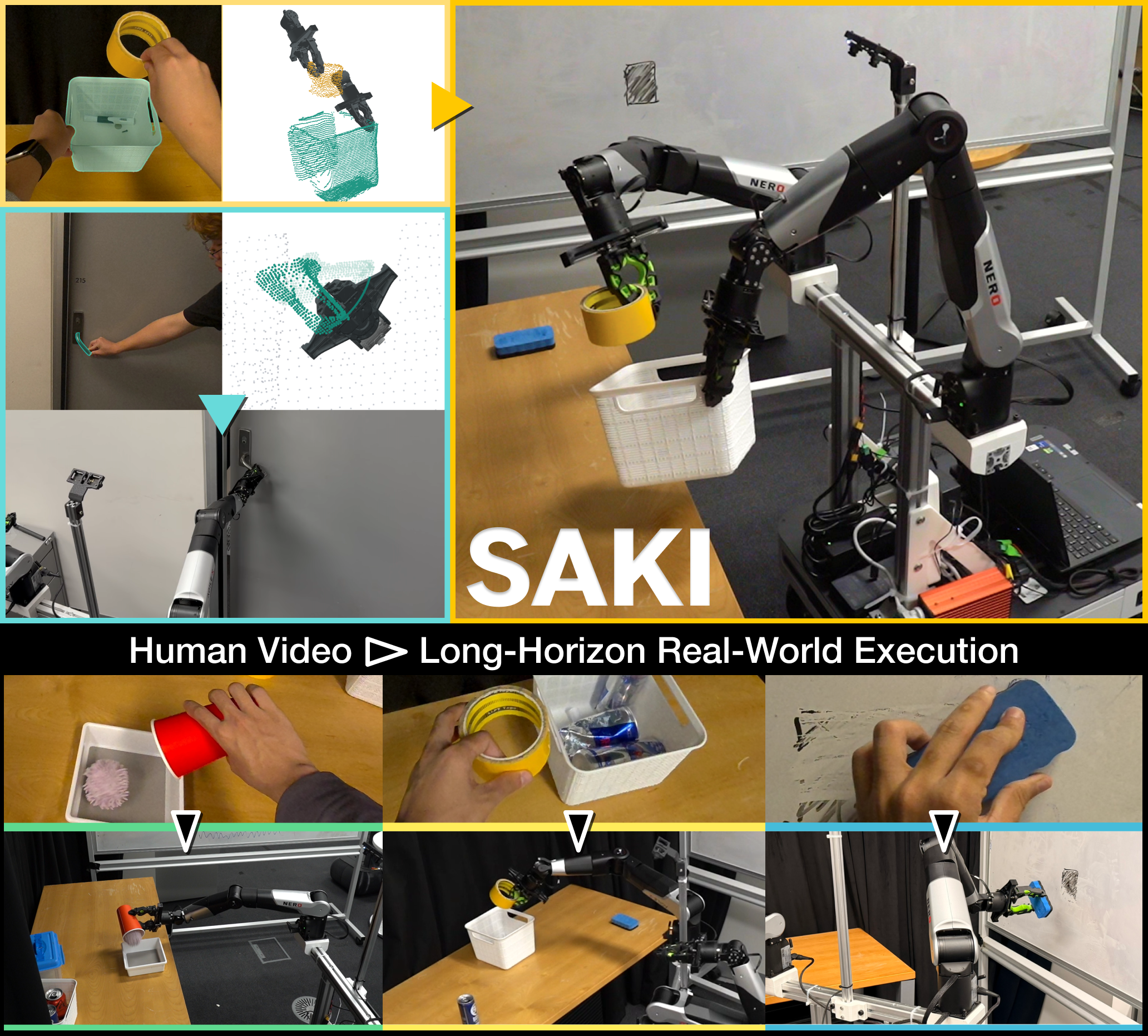}
\caption{SAKI transfers object motion and contact roles across articulation, pouring, placement and wiping. Separate task examples include bimanual placement, where container support accompanies manipulation.}
\label{fig:teaser}
\end{figure}

\section{Introduction}

Consider a mobile robot asked to tidy a workspace: dispose of a can, place tape in a basket, and wipe a whiteboard. The interactions may be demonstrated separately, at different locations and from different viewpoints. Executing them together requires the robot to preserve each interaction while adapting its motion to the objects it currently observes and the configuration reached by the preceding task. The same demonstrated skill must therefore remain usable both in a new layout and partway through a mobile sequence.

\begin{figure*}[t]
\centering
\includegraphics[width=\linewidth]{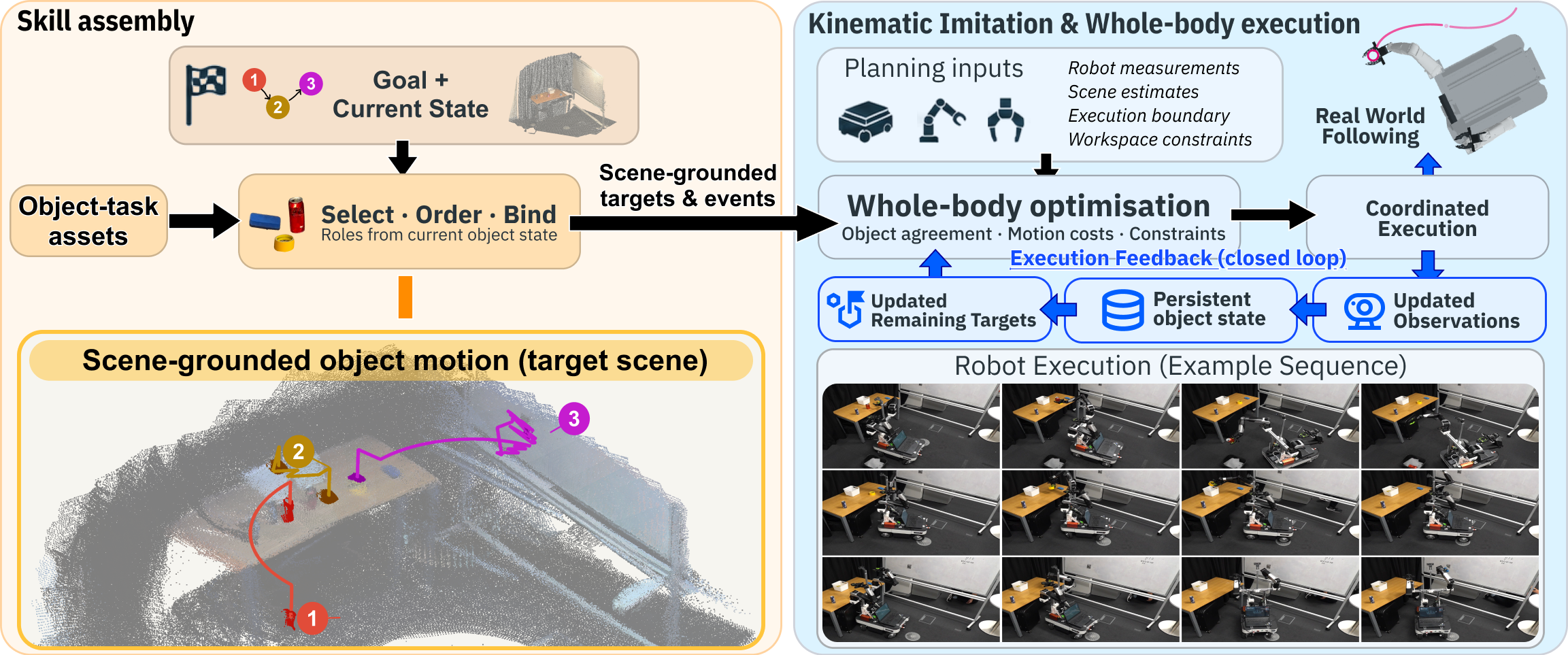}
\caption{SAKI skill assembly and closed-loop execution. Object-role bindings ground reusable skills in the current scene, and whole-body optimisation generates coordinated robot motion. Accepted visual updates revise the remaining targets while replacement plans preserve committed motion. Paired base/arm completion and gripper events advance the sequence; retained scene estimates and the preceding terminal robot configuration initialise the next approach.}
\label{fig:overview}
\end{figure*}

Human videos can provide the object motion needed for these interactions. Object-centric imitation can transfer demonstrated motion across arrangements~\cite{heppert2024ditto} or recover manipulation plans~\cite{zhu2024orion}. However, video demonstrations interweave essential interaction geometry with motion specific to the demonstrator and the original scene; a recovered trajectory alone does not distinguish between them. Placement requires clearance and terminal alignment, but the transfer path can change with the layout. Closely reproducing the demonstrated path can retain unnecessary detours or constrain the robot's base approach. During mobile execution, the robot's observations change, and earlier actions alter the scene and robot configuration from which subsequent interactions begin. Dynamic memory and skill-based systems address scene persistence and task execution~\cite{liu2024dynamem,shah2024bumble}. For imitation, this information must determine motion at two timescales: observations update targets within a skill, and preceding actions establish the geometry and robot state for its successor. A destination must remain available even when turning towards the next object takes it out of view.

We propose SAKI (Skill Assembly and Kinematic Imitation), a framework for reusing demonstrated interactions through state-dependent motion generation. SAKI preserves the requirements of a demonstrated interaction while regenerating the robot motion needed to realise it in the current scene. Object-task assets retain demonstrated contact, clearance, orientation and terminal relations, together with rules permitting transfer motion to change. Object-role bindings locate these requirements in the current scene, and whole-body optimisation finds a coordinated base-arm realisation. Interaction-reference preparation determines the object motion to be realised, while whole-body optimisation determines how the robot realises it.

The framework uses distinct runtime information for grounding and continuation. Retained object estimates locate interaction targets across views; measured base, arm and gripper states describe the robot configuration; execution context identifies the current skill and the motion already committed to the robot. An accepted observation can revise future targets while the replacement plan joins the retained execution boundary. At a skill switch, the preceding terminal configuration initialises the next approach. These interfaces let the tidying sequence connect independently prepared interactions while carrying object estimates and robot configuration forward through execution.

\noindent\textbf{The paper makes three contributions:}

\begin{itemize}
\item \textbf{The SAKI framework for reusable demonstrated interactions.} We separate retained interaction requirements from scene-dependent robot motion, combining object-task assets, role binding and whole-body kinematic imitation to realise independently demonstrated interactions in mobile tasks.
\item \textbf{Cross-view closed-loop kinematic imitation.} Persistent object estimates retain task geometry across views, and accepted visual updates regenerate remaining targets and coordinated robot motion. Replacement plans join the retained execution boundary, preserving commands already committed to the robot.
\item \textbf{Cross-demonstration skill assembly.} Selection, repetition, ordering and object-role binding compose independently prepared assets under supplied goals and dependencies. Retained scene estimates and the preceding terminal robot configuration initialise successor motion, connecting separate demonstrations into continuous mobile tasks.
\end{itemize}

\section{Related Work}

\textbf{Object-Centric Imitation and Interaction Geometry:} DITTO transfers object-relative motion~\cite{heppert2024ditto}, while ORION recovers object graphs~\cite{zhu2024orion}. SPOT conditions a policy on target-relative pose trajectories~\cite{hsu2024spot}; Robot See Robot Do~\cite{kerr2025rsrd} and ScrewMimic~\cite{bahety2024screwmimic} use articulated-part trajectories and screw motion for bimanual planning. JFTO jointly optimises grasps and robot motion against a learned trajectory distribution and collision constraints~\cite{dong2026jfto}. NDF~\cite{simeonov2021ndf} and relational NDF~\cite{simeonov2022rndf} transfer manipulation poses and object-part alignments. VoxPoser~\cite{huang2023voxposer}, ReKep~\cite{huang2024rekep} and CrossInstruct~\cite{barron2026crossinstruct} express task geometry through spatial value maps, keypoint constraints and 3D motion distributions, respectively. CrossInstruct derives these distributions from visual instructions, VLM reasoning and multi-view pointing. SAKI couples recovered object motion to phase-specific requirements that determine which segments can change while retaining the demonstrated interaction.

\textbf{Mobile Manipulation and Execution Feedback:} Mobile ALOHA learns from whole-body teleoperation~\cite{fu2024mobilealoha} and HoMMI from human demonstrations~\cite{xu2026hommi}. HRT1 transfers human-derived trajectories through base positioning and arm-trajectory optimisation~\cite{allu2025hrt1}. Perceptive MPC addresses continuous mobile manipulation~\cite{pankert2020perceptivempc}, while ActPerMoMa plans motion using information gain and grasp reachability~\cite{jauhri2024actpermoma}. Inner Monologue feeds scene and outcome information into planning~\cite{huang2022innermonologue}; DynaMem~\cite{liu2024dynamem} and BUMBLE~\cite{shah2024bumble} maintain scene information for mobile tasks. In SAKI, retained object estimates locate interaction targets across views. Accepted visual updates revise those targets, and whole-body replanning connects the remaining motion to the retained execution boundary.

\textbf{Demonstration Reuse and Skill Composition:} MimicGen adapts demonstrations to new contexts~\cite{mandlekar2023mimicgen}, and DemoGen transforms trajectories and point clouds for new object configurations~\cite{xue2025demogen}. TriManPolicy retimes sensorimotor segments under task dependencies and arm-usage constraints~\cite{zhao2026trimanpolicy}; WANDA connects contact-rich segments through whole-body planning to generate training data~\cite{guo2026wanda}. Dream2Flow recovers object flow from generated video~\cite{dharmarajan2025dream2flow}, while RoboReact combines generated video, interaction reconstruction, refinement and online re-grounding~\cite{he2026roboreact}. XSkill composes learned skill prototypes~\cite{xu2023xskill}, SayCan grounds skill selection with value functions~\cite{ahn2022saycan}, and Code as Policies combines perception and control APIs~\cite{liang2022codeaspolicies}. SAKI connects independently demonstrated interactions through object-role bindings, with retained scene estimates and the preceding terminal robot configuration initialising each successor. Prepared references support execution and reuse across layouts.

\section{Method}

SAKI prepares reusable skills from human demonstrations, binds them to objects in the execution scene, and generates whole-body motion (Fig.~\ref{fig:overview}). During execution, visual updates revise the remaining motion, and the preceding terminal configuration initialises each successor skill.

\textbf{State used for motion generation.} SAKI maintains scene estimates, measured robot state and execution context. Scene estimates contain object identities, geometry, poses and observation records in a common frame; they locate the objects used to generate interaction targets. Measured robot state contains the base pose, arm joint positions and gripper state, providing the initial configuration and execution feedback. Execution context records the active skill invocation, progress, accepted plan and committed motion prefix. When a plan is replaced, this context supplies a continuation boundary on the accepted plan. The latest robot measurement describes the current physical configuration.

\begin{figure}[t]
\centering
\includegraphics[width=\linewidth]{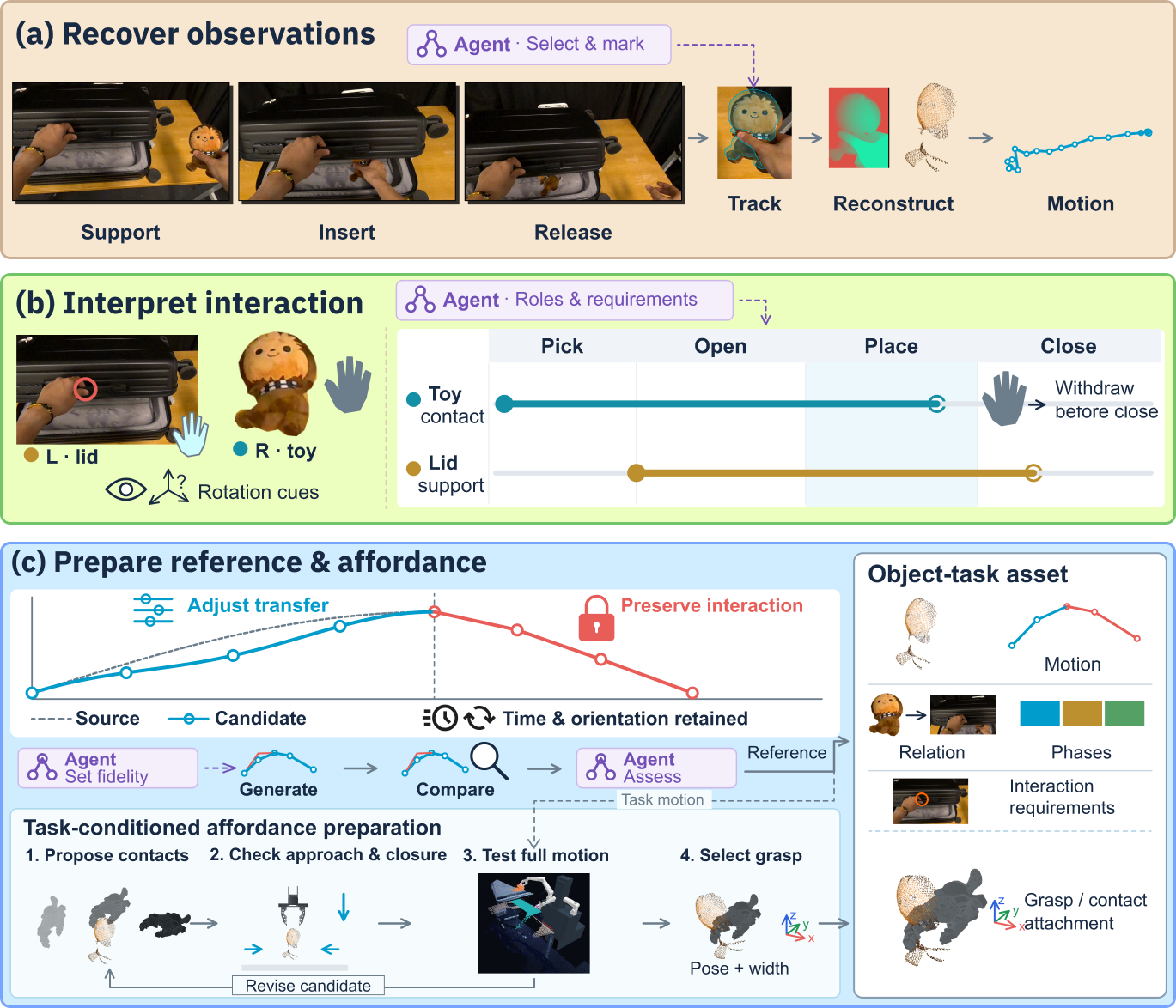}
\caption{Skill preparation separates interaction requirements from adjustable motion. Object roles, phases and contact requirements guide reference and grasp preparation for scene-dependent execution. Curves and contacts are schematic; the inset shows kinematic replay.}
\label{fig:mechanism}
\end{figure}

\subsection{Video-Derived Skill Assets}\label{method:skills}

\textbf{Skill representation.} An object-task asset packages an accepted object-motion reference, object roles, ordered phases, contact events and adaptation rules. Manipulated, reference and supporting roles identify the participants in an interaction; phases delimit intervals with different motion requirements. An invocation assigns these roles to scene entities, retaining the specification while regenerating its targets and robot realisation.

\textbf{Recover object motion.} FoundationStereo estimates depth from rectified stereo images~\cite{wen2025foundationstereo}, and SAM 2 supplies tracked object masks~\cite{ravi2024sam2}. Calibration and camera poses register timestamped object clouds in the source world. The reference retains observed positions and supported orientation components; unobservable rotations remain unspecified. For rigid transport, the selected interval ends before release. Connections inferred across missing observations are distinguished from measured motion.

\textbf{Prepare an interaction reference.} Before execution, a preparation agent reviews video, recovered motion and the supplied task goal to propose roles, phase boundaries, contact events and permitted reference changes. Each phase specifies position/orientation requirements and a motion rule. Adjustable segments provide control points and an interpolation law; protected intervals retain lift clearance, pouring orientation or wiping contact. Compatible geometry and contact models are preparation inputs. The agent authors and reviews proposals, while deterministic tools generate and check candidate references. The preparation instance uses Codex with GPT-6 and saves phase assessments, object-pose keyframes and base/arm initialisation for compilation.

Candidates retain protected source samples, source timing and available orientation observations, using linear or piecewise cubic Hermite interpolation (PCHIP) on adjustable position segments. Each segment records its source interval, position and orientation requirements, permitted changes and assessment. The agent reviews position deviations and phase junctions in sampled source/candidate comparisons; acceptance requires an assessment for every declared segment. Accepted references retain their source observations and assessments for scene binding. Robot-specific timing is determined after motion solving.

\textbf{Connect requirements to motion.} The phase record links preparation to compilation: its source interval identifies reference samples, its motion rule selects preserved or interpolated positions, and its contact events delimit attachment and release. A prepared grasp converts object motion into end-effector targets. Textual requirements guide reference assessment, while the interaction model and solver configuration supply numerical enforcement.

For can disposal, a 24-frame reference separates lift, transfer and final alignment. Its lift endpoint and final-interaction boundary define the adaptation interval in Section~\ref{method:adaptation}. The transfer admits positional correction between these boundaries, while the terminal event schedules gripper opening after alignment. Fig.~\ref{fig:mechanism} illustrates preparation for bimanual packing, where support must persist during the other arm's placement.

\subsection{Skill Assembly and Scene Grounding}\label{method:adaptation}

\textbf{Assemble compatible interactions.} Given a goal and supplied dependencies, the agent selects and orders existing skills. Each invocation names an asset, binds its object roles to scene entities and specifies terminal gripper intent. Repetition creates another invocation with new bindings. Compatibility preserves attachment, support and release events: transport retains the held object, and insertion retains the supporting contact. The compiler combines this invocation specification with scene estimates and the initial robot configuration to generate grounded targets and connecting approaches.

\textbf{Bind roles to the scene.} Calibrated observations and capture-time odometry register scene entities in a common execution frame; a base scan acquires entities across viewpoints when needed. Each invocation binds its manipulated, reference and supporting roles to these entities. Their retained estimates provide the anchors for target generation even when an entity leaves the current image.

The compiler generates object targets from the accepted reference and current scene anchors. Approach motion uses the source object and prepared grasp, final alignment uses the destination, and transfer connects these regions. The reference and phase requirements remain fixed across invocations; object targets, robot motion and execution timing are regenerated for the current scene.

\begin{figure}[t]
\centering
\includegraphics[width=\linewidth]{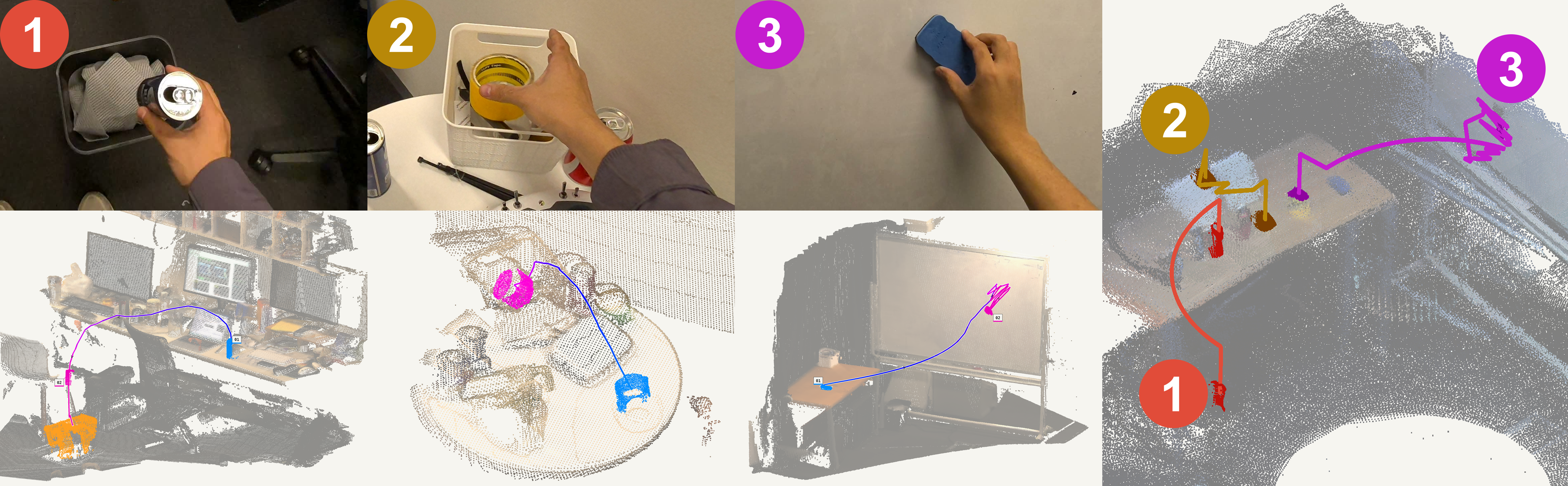}
\caption{Skill assembly across demonstration scenes. Object-role bindings place prepared interaction references in a shared target scene, while the selected order and successor approaches connect them into a mobile task.}
\label{fig:demonstration-interactions}
\end{figure}

\textbf{Example: tidying and wiping (L1).} Three independently prepared assets serve can disposal, tape placement and board wiping (Fig.~\ref{fig:demonstration-interactions}; Table~\ref{tab:assembly}). Assembly assigns their order, scene-object bindings and terminal events. Container openings determine release targets; the board frame determines tool orientation and sweeping motion. These bindings provide the input to motion generation, while Section~\ref{method:execution} explains how execution connects the invocations.

\begin{table}[t]
\centering
\footnotesize
\setlength{\tabcolsep}{3pt}
\renewcommand{\arraystretch}{1.12}
\begin{tabularx}{\linewidth}{@{}>{\raggedright\arraybackslash}p{.20\linewidth}>{\raggedright\arraybackslash}X>{\raggedright\arraybackslash}X@{}}
\toprule
Prepared skill & Retained interaction & L1 invocation \\
\midrule
Can disposal & Lift, transport, final alignment and release & Can to floor bin; terminal release \\
Tape placement & Lift, align with opening and release & Tape to tabletop basket; terminal release \\
Board wiping & Tool grasp, surface contact and sweeping & Eraser to marked board region; terminal hold \\
\bottomrule
\end{tabularx}
\caption{Three independently prepared skills retain their interaction requirements while binding to objects in one tidying-and-wiping sequence.}
\label{tab:assembly}
\end{table}

\textbf{Confine endpoint correction to transfer.} For the rigid-transport path, let \(\bar p(s)\) be the accepted object centre curve at task progress \(s\in[0,1]\), and let \(a\) and \(g\) be the current contact and release anchors. The linear map \(B\) aligns and scales horizontal displacement to the new layout while retaining metric vertical displacement. The remaining endpoint mismatch is \(\Delta=g-a-B(\bar p(1)-\bar p(0))\). The adapted path is \(p^*(s)=a+B(\bar p(s)-\bar p(0))+w(s)\Delta\).

The accepted lift end \(s_L\) and final-interaction start \(s_I\) satisfy \(0\leq s_L<s_I\leq1\). We use \(u(s)=\operatorname{clip}((s-s_L)/(s_I-s_L),0,1)\) and \(w(s)=3u(s)^2-2u(s)^3\). Correction is zero through the lift, changes over transfer, and is constant throughout the final interaction. Thus \(p^*(0)=a\) and \(p^*(1)=g\), while the mapped protected segments retain their relative displacements. During initial scene binding, a changed destination height can be absorbed before terminal lowering begins.

In can disposal, the observed can pose and bin opening supply the anchors, and the prepared grasp converts the adapted object path into tool centre point (TCP) targets. Initial binding can adapt destination height; online corrections use the horizontal update described in Section~\ref{method:execution}. The mapping assumes nondegenerate horizontal separation and an admissible transfer interval.

\subsection{Whole-Body Kinematic Imitation}\label{method:imitation}

Whole-body optimisation turns grounded object targets into coordinated base and arm motion. Candidate approaches are evaluated against the complete attached trajectory, since reaching the grasp alone does not ensure that the following interaction is feasible. Initial motion starts from measured robot state; a replacement plan joins the boundary supplied by execution context. Stored solutions serve as numerical initialisation.

Let \(z_k=(b_k,q_k)\) contain planar base pose \(b_k\in SE(2)\) and arm joints \(q_k\) at sample \(k\), and let \(\tau_k^*\) contain the grounded object targets. The objective combines agreement with those targets and motion regularisation:

\begin{equation}\label{eq:whole-body}
\begin{aligned}
\min_{z_{0:K}}\quad &\sum_{k=0}^{K}\ell_{\mathrm{object}}(z_k,\tau_k^*)+\mathcal R(z_{0:K}),\\
\ell_{P,k} &= \frac{1}{P}\sum_{l=1}^{P}\|\hat p_{k,l}-p^*_{k,l}\|^2.
\end{aligned}
\end{equation}

For rigid transport, the point residual compares target points \(p^*_{k,l}\) with predictions \(\hat p_{k,l}=F_E(z_k)\,{}^{E}T_O\,p_l^O\). Here \(F_E\) is end-effector forward kinematics, \({}^{E}T_O\) is the prepared object-to-end-effector attachment transform, and \(p_l^O\) is an object-frame point; homogeneous coordinates are implicit. This residual contributes to \(\ell_{\mathrm{object}}\), with pose terms using supported observations. An observed axis and the grasp convention supply distinct orientation information.

The regulariser \(\mathcal R\) combines smoothness, base-motion, posture and configured clearance costs. Joint bounds, base-motion constraints and fixed execution boundaries restrict the solve. Accepted base and arm trajectories are retimed under motion limits and paired with phase-specific gripper events. The interaction model supplies the robot-object relation; the point residual above instantiates rigid attachment.

\textbf{Represent contact-dependent interactions.} For arm \(a\), the target \(T^*_{W,E_a}(t)=T_{W,R}T_{R,O_a}(t)T_{O_a,E_a}\) combines scene registration, object motion relative to a reference entity and the prepared grasp. In bimanual packing, the supporting arm retains its lid-contact target and closed-gripper intent throughout the other arm's insertion and release interval. Both arms are solved on a common time base, so support persists across the placement phase. Task-specific preparation supplies the contact geometry and schedule; target compilation, solving and execution are shared.

\textbf{Pouring and wiping through pose targets.} During pouring, the cup-to-gripper attachment remains fixed while the cup orientation changes relative to the receiving vessel, then returns upright. This rotation enters the object-pose target at each sample and is transferred to the wrist through the prepared grasp. Wiping instead uses a board frame, marked region and tool-to-contact transform, with a grasp that leaves the wiping face unobstructed. The strokes align the felt normal with the board normal and retain the prepared contact offset while in-plane coordinates trace the sweep. Engagement and withdrawal decrease and increase the normal separation, respectively. The gripper remains closed through wiping; L1 also retains the tool after withdrawal. Both interactions use the same pose-target construction: scene registration locates the reference entity, the phase specifies object motion, and the grasp transform determines the end-effector target. Fig.~\ref{fig:wiping-outcome} presents their physical execution.

\textbf{Observation opportunities.} Adjustable transfer permits changes in approach and camera heading; optional surface-view costs favour useful viewing geometry. As the base moves, a destination can become visible from a closer or less occluded viewpoint. The new observation is registered into persistent object state; an accepted estimate then revises future targets through Section~\ref{method:execution}.

\subsection{Stateful Closed-Loop Execution}\label{method:execution}

\textbf{Update motion within an invocation.} New observations revise the scene estimates used by the active invocation. Consistent observations update object pose; occlusion or ambiguous identity preserves the retained estimate, while attached objects propagate through robot kinematics. Observation and entity revisions associate each solve with the estimates it consumed. Accepted revisions change unexecuted targets with participant identities and the prepared reference fixed. For can disposal, online corrections change horizontal destination-relative translation, retaining the prepared height and orientation.

The next permitted solve consumes the accepted scene snapshot and the continuation boundary from execution context. It regenerates and jointly retimes the remaining base, arm and gripper motion while retaining the committed prefix. Thus a target revision changes future motion without replacing commands already issued to the robot. Candidate promotion records the accepted executable, and robot reports confirm activation. Subsequent planning carries that executable as its source plan. Fig.~\ref{fig:closed-loop-update} traces the estimate-to-target-to-plan connection in a recorded run.

\textbf{Carry state between invocations.} Sequence preparation initialises each approach from the preceding terminal base pose, arm joints and gripper state, retaining scene-object identities and estimates. During execution, paired arm/base completion and gripper events govern progression. If a visual correction changes the current task's endpoint, the sequence retargeter rebuilds the immediately following approach from the revised terminal configuration; later references remain available for their own binding. The next invocation therefore starts from the preceding task's result while using its own interaction requirements and bound objects.

In L1, the tape approach starts from the base, arm and open-gripper state left by can disposal. After tape release, the robot approaches the eraser and retains its grasp through wiping. Persistent estimates preserve scene context across these viewpoint changes. Table~\ref{tab:capability} evaluates whether the assembled sequence reaches all three physical outcomes without intermediate resets or assistance.

\section{Experiments}

The evaluation examines how the SAKI framework reuses interaction specifications: continuous tasks test their composition, reference ablation tests the benefit of preparation, and changed-layout trials test reuse with frozen skills. A recorded visual update follows the runtime connection from object estimates to activated motion. Local manipulation comparisons evaluate reference-generation pipelines under shared downstream conditions.

\subsection{Setup and Shared Protocol}\label{exp:protocol}

\textbf{Platform and tasks.} We use a Ranger Mini V3 with two Nero arms. Skills are prepared from calibrated stereo human demonstrations and task context with compatible grasp or contact information. Seven single tasks and three continuous sequences each have 15 initiated trials. A separate local suite tests tape placement into a stationary basket, same-table pouring and a reachable wiping region. These operations assess interaction execution within fixed-base reach; the complete suite additionally requires mobile realisation across the workspace.

\textbf{Success and trial accounting.} Success requires the complete specified outcome. S1 requires the door to reach its fully open position; wiping requires removal of at least 80\% of the original markings in the designated region, judged by a human reviewer from the execution video. This is a visual outcome assessment rather than a pixel-area measurement. Local placement includes release into the basket, and local pouring includes transferring the ball into the tray, returning the cup upright and placing it down. For continuous sequences, every evaluated stage must complete without intermediate resets or human assistance. A trial that completes an early manipulation but fails a later stage is unsuccessful for the full sequence.

Complete-task rates include all 15 initiated attempts. Stage rates use the attempts reaching that stage, localising losses without changing the full-sequence denominator. Matched comparisons retain common layout conditions, success definitions and execution budgets, including planning failures and timeouts. Videos and motion records illustrate execution; trial outcomes determine success.

\subsection{Complete-Pipeline Capability}

Table~\ref{tab:capability} defines seven single tasks and three continuous sequences. \textbf{L1, tidying and wiping}, executes the three skills assembled in Section~\ref{method:adaptation}. \textbf{L2, collection and pouring}, loads balls into a cup and pours them into a kettle. \textbf{L3, three-location collection}, acquires a basket and can at A, deposits the can, then collects further cans at B and C. Their three, two and three evaluated stages can each contain multiple skill invocations.

\begin{table}[t]
\centering
\footnotesize
\setlength{\tabcolsep}{3pt}
\renewcommand{\arraystretch}{1.12}
\begin{tabularx}{\linewidth}{@{}>{\raggedright\arraybackslash}Xr@{}}
\toprule
Task & SAKI n/N (\%) \\
\midrule
\multicolumn{2}{@{}l}{\emph{Single tasks}}\\
S1 Door fully open & 7/15 (46.7\%) \\
S2 Long whiteboard wiping & 11/15 (73.3\%) \\
S3 Basket handling and tape placement & 13/15 (86.7\%) \\
S4 Can disposal & 15/15 (100.0\%) \\
S5 Pouring & 15/15 (100.0\%) \\
S6 Lid opening and object placement & 14/15 (93.3\%) \\
S7 Box opening and object placement & 10/15 (66.7\%) \\
\midrule
\multicolumn{2}{@{}l}{\emph{Continuous sequences}}\\
L1 Tidying + wiping & 9/15 (60.0\%) \\
L2 Collection + pouring & 13/15 (86.7\%) \\
L3 Three-location collection & 9/15 (60.0\%) \\

\bottomrule
\end{tabularx}
\caption{Complete physical outcomes across seven single tasks and three continuous sequences, with 15 initiated trials per condition. Sequence success requires every stage to complete.}
\label{tab:capability}
\end{table}

\begin{figure}[t]
\centering
\includegraphics[width=0.95\linewidth]{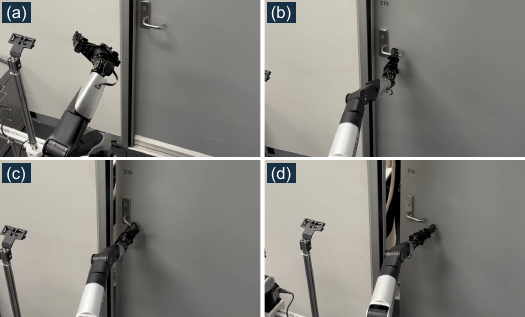}
\caption{Door opening couples handle engagement to articulated motion. Contact is maintained as the door transitions to partial opening, illustrating an interaction that extends beyond the initial grasp.}
\label{fig:door-interaction}
\end{figure}

\begin{figure}[t]
\centering
\includegraphics[width=0.95\linewidth]{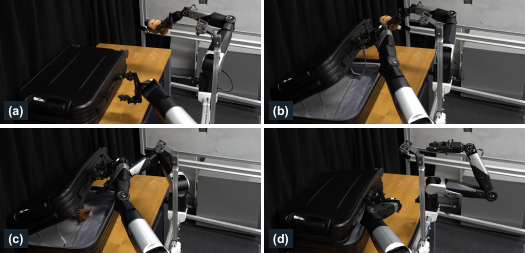}
\caption{Autonomous suitcase packing requires lid support to overlap with toy placement. One arm maintains the opening while the other releases the toy and withdraws before closure, illustrating concurrent contact roles.}
\label{fig:suitcase-packing}
\end{figure}

\begin{figure*}[t]
\centering
\includegraphics[width=\linewidth]{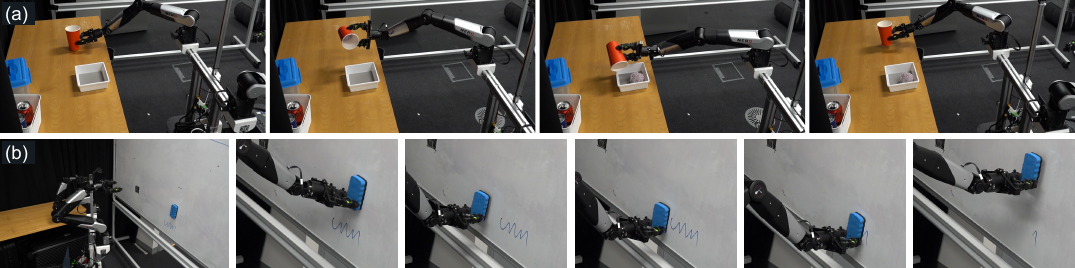}
\caption{Phase-specific pose requirements govern physical interaction: time-varying cup orientation during pouring (a) and sustained tool alignment during wiping (b).}
\label{fig:wiping-outcome}
\end{figure*}

\textbf{Complete-task outcomes.} SAKI completes 85/105 single-task trials (81.0\%) and 31/45 continuous-sequence trials (68.9\%). Can disposal and pouring succeed in all single-task attempts; lower completion for door opening, long wiping and box opening identifies extended interaction as a remaining challenge.

\textbf{Assembly through physical execution.} All 15 L1 attempts complete can disposal and tape placement; nine also complete wiping. The six full-sequence losses therefore occur at the final stage, after the initial skills have been connected successfully. L1 provides physical evidence that independently prepared assets can be bound and executed as one mobile sequence. L2 has conditional stage rates of 14/15 and 13/14, losing one attempt in collection and one in pouring. L3 achieves 13/15, 11/13 and 9/11 at A, B and C; its first stage includes basket acquisition, can acquisition and placement. These outcomes distinguish L1's concentrated final-stage loss from losses accumulated across L3.

\textbf{Interaction requirements.} The qualitative cases expose complementary demands on the motion representation: articulated engagement (Fig.~\ref{fig:door-interaction}), overlapping support and placement (Fig.~\ref{fig:suitcase-packing}), and phase-dependent orientation and contact (Fig.~\ref{fig:wiping-outcome}). Suitcase packing requires support throughout the other arm's insertion and withdrawal. Pouring requires a changing cup orientation with a retained grasp; wiping requires sustained tool alignment as contact moves across the board. These requirements enter the pose targets and contact schedules of Section~\ref{method:imitation}, extending the interaction specification beyond object-centre displacement.

\subsection{Reference Optimisation on Continuous Sequences}

\begin{table}[t]
\centering
\footnotesize
\setlength{\tabcolsep}{3pt}
\renewcommand{\arraystretch}{1.12}
\begin{tabularx}{\linewidth}{@{}>{\raggedright\arraybackslash}Xrr@{}}
\toprule
Sequence & \shortstack{SAKI w/o\\ref. optimisation} & SAKI \\
\midrule
L1 Tidying + wiping & 2/15 (13.3\%) & 9/15 (60.0\%) \\
L2 Collection + pouring & 4/15 (26.7\%) & 13/15 (86.7\%) \\
L3 Three-location collection & 1/15 (6.7\%) & 9/15 (60.0\%) \\
\bottomrule
\end{tabularx}
\caption{Reference preparation improves continuous-task completion with scene grounding, whole-body solving and online updates retained.}
\label{tab:ref-ablation}
\end{table}

Table~\ref{tab:ref-ablation} replaces task-conditioned reference refinement with direct interpolation of the reconstructed object trajectory. Scene binding, whole-body optimisation, visual updates and replanning remain active. The solver can change the base-arm realisation but must follow the supplied object path, isolating the benefit of reference preparation within this downstream setup. Completion increases from 7/45 (15.6\%) to 31/45 (68.9\%), a gain of 53.3 percentage points, with improvement on every sequence. This shows that retaining whole-body solving and feedback does not remove the benefit of preparing the interaction reference. For L1, the same downstream mechanisms complete 2/15 sequences with direct interpolation and 9/15 with preparation. An interpolated object path can restrict base rotation and posture changes that improve reach or target observation. Failed ablated trials included poor target observation and collisions with task objects. Allowing transfer changes gives the downstream solve freedom to accommodate both observation and motion feasibility while preserving the interaction. The completion comparison measures the resulting benefit across the full sequence.

\begin{figure}[t]
\centering
\includegraphics[width=\linewidth]{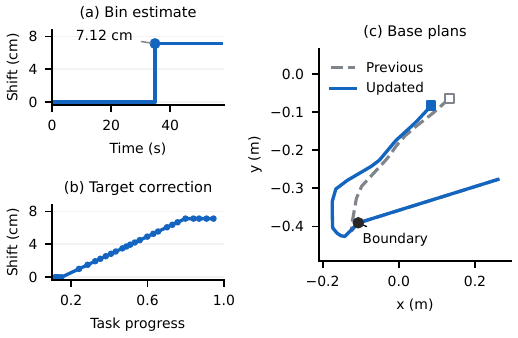}
\caption{Visual feedback revises remaining motion while preserving committed execution. A 7.12 cm shift in the retained bin estimate (a) modifies reference targets (b) and the accepted base plan (c); execution reports confirm activation.}
\label{fig:closed-loop-update}
\end{figure}

\begin{table}[t]
\centering
\footnotesize
\setlength{\tabcolsep}{3pt}
\renewcommand{\arraystretch}{1.12}
\begin{tabularx}{\linewidth}{@{}>{\raggedright\arraybackslash}Xrr@{}}
\toprule
Shared subtask & DITTO-adapted & SAKI \\
\midrule
Tape into stationary basket & 13/15 (86.7\%) & 15/15 (100.0\%) \\
Same-table cup pouring & 10/15 (66.7\%) & 15/15 (100.0\%) \\
Local whiteboard wiping & 8/15 (53.3\%) & 15/15 (100.0\%) \\
\bottomrule
\end{tabularx}
\caption{Local manipulation under shared perception, grasping and execution modules, with 15 trials per operation within fixed-base reach.}
\label{tab:ditto}
\end{table}

\begin{figure}[t]
\centering
\includegraphics[width=\linewidth]{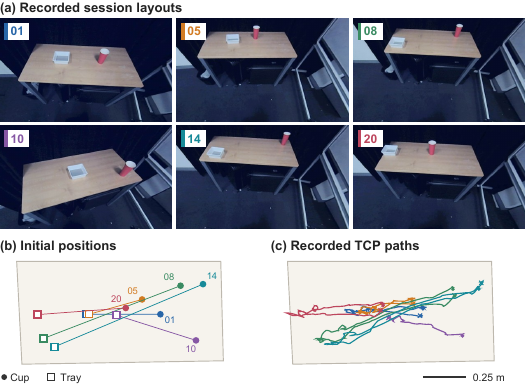}
\caption{A frozen pouring skill accommodates changed cup, tray and robot starting configurations. Six of twenty trials pair initial arrangements with feedback-derived TCP paths at a common metric scale, registered to the fixed table.}
\label{fig:pouring-generalisation}

\vspace{4pt}
\centering
\footnotesize
\setlength{\tabcolsep}{3pt}
\renewcommand{\arraystretch}{1.12}
\begin{tabularx}{\linewidth}{@{}>{\raggedright\arraybackslash}p{89pt}>{\raggedright\arraybackslash}Xr@{}}
\toprule
Skill & Changed scene condition & \shortstack{Success\\n/N (\%)} \\
\midrule
S4 Can disposal & Source + destination & 18/20 (90.0\%) \\
S5 Pouring & Cup + tray + robot start & 20/20 (100.0\%) \\
S6 Lid opening + placement & Source + destination + robot start & 15/20 (75.0\%) \\
\bottomrule
\end{tabularx}
\captionof{table}{Frozen-skill reuse under changed scene conditions. References and interaction requirements remain fixed; bindings and robot motion are recomputed.}
\label{tab:asset-transfer}
\end{figure}

\subsection{Visual Updates in Recorded Execution}\label{exp:closed-loop}

During a recorded can-disposal execution, an accepted visual update shifts the estimated bin position by 7.12 cm horizontally (Fig.~\ref{fig:closed-loop-update}a). The next progress-triggered solve uses this estimate to revise the remaining object targets. Their orientations remain unchanged, while the later position corrections progressively reach the estimated displacement (Fig.~\ref{fig:closed-loop-update}b). The replacement plan retains the executable points through the continuation boundary and changes the subsequent base and arm motion and its timing (Fig.~\ref{fig:closed-loop-update}c). Robot reports confirm that this plan was activated, and the next planning request uses it as the source plan. The update therefore changes the motion used during execution while retaining the same prepared skill and bound objects.

\subsection{Comparison on Shared Local Tasks}

The local comparison tests reference-generation pipelines across placement, orientation-changing pouring and sustained wiping contact within fixed-base reach. DITTO-adapted integrates DITTO's object-trajectory extraction and transformation with our robot (Table~\ref{tab:ditto})~\cite{heppert2024ditto}. The methods share perception, grasping and execution modules, with method-specific reference generation and identical local outcome definitions. Across the three equally sampled operations, SAKI achieves 45/45 completions (100.0\%) versus 31/45 (68.9\%) for DITTO-adapted. The larger gaps in pouring and wiping associate the improvement with operations requiring orientation change or sustained contact. The complete tasks additionally test long wiping over a larger region and basket handling with support and return around placement. Together, the two suites assess local interaction quality and completion of the surrounding mobile task.

\subsection{Position Generalisation and Skill Reuse}

Table~\ref{tab:asset-transfer} tests reuse with the accepted reference, phase boundaries and interaction requirements frozen. Each trial regenerates bindings, targets and plans for changed scene conditions. In S6, source and destination positions change and the robot start is randomised among locations with both objects visible; execution includes the initial mobile approach. Pouring varies cup and tray positions and robot starting configuration while keeping the table fixed. All twenty trials complete acquisition, pouring, upright recovery and placement. Fig.~\ref{fig:pouring-generalisation} links six initial arrangements to TCP paths reconstructed from measured joints and base odometry, from grasp to release. The paths share a metric scale and fixed-table registration, making the changed robot motion comparable across layouts. Can disposal completes 18/20 trials (90.0\%), and lid opening with placement completes 15/20 (75.0\%). Together with pouring, these results establish position reuse across three interactions while keeping their prepared specifications fixed. L1 tests the complementary capability of connecting different specifications into one task.

\section{Conclusion and Future Work}

SAKI provides a framework for carrying human-demonstrated interactions across changing scenes and continuous mobile execution. Retained interaction requirements define what the robot must realise; object-role bindings, scene estimates and execution boundaries determine its current motion. Real-robot results establish the benefit of reference preparation with whole-body solving and feedback retained, and demonstrate continuous composition and frozen-skill reuse. The recorded visual update connects a revised estimate to changed targets and an activated replacement plan while preserving committed motion. Together, these findings support the framework's separation of reusable interaction specifications from the robot trajectories that ground, update and connect them.
Building on SAKI's demonstrated ability to reuse and compose interactions across changing scenes, future work will extend this capability to more complex task dependencies, incorporating richer scene information and execution feedback to support broader and more complex task-level reasoning and adaptation.
\bibliographystyle{IEEEtran}
\bibliography{refs}

@inproceedings{heppert2024ditto,
  author = {Nick Heppert and Max Argus and Tim Welschehold and Thomas Brox and Abhinav Valada},
  title = {{DITTO}: Demonstration Imitation by Trajectory Transformation},
  booktitle = {IEEE/RSJ International Conference on Intelligent Robots and Systems (IROS)},
  year = {2024},
  doi = {10.1109/IROS58592.2024.10801982}
}

@article{allu2025hrt1,
  author = {Sai Haneesh Allu and Jishnu Jaykumar P and Ninad Khargonkar and Tyler Summers and Jian Yao and Yu Xiang},
  title = {{HRT1}: One-Shot Human-to-Robot Trajectory Transfer for Mobile Manipulation},
  journal = {arXiv preprint arXiv:2510.21026},
  year = {2025}
}

@article{zhu2024orion,
  author = {Yifeng Zhu and Arisrei Lim and Peter Stone and Yuke Zhu},
  title = {Vision-based Manipulation from Single Human Video with Open-World Object Graphs},
  journal = {arXiv preprint arXiv:2405.20321},
  year = {2024}
}

@article{hsu2024spot,
  author = {Cheng-Chun Hsu and Bowen Wen and Jie Xu and Yashraj Narang and Xiaolong Wang and Yuke Zhu and Joydeep Biswas and Stan Birchfield},
  title = {{SPOT}: {SE(3)} Pose Trajectory Diffusion for Object-Centric Manipulation},
  journal = {arXiv preprint arXiv:2411.00965},
  year = {2024}
}

@article{xu2026hommi,
  author = {Xiaomeng Xu and Jisang Park and Han Zhang and Eric Cousineau and Aditya Bhat and Jose Barreiros and Dian Wang and Jeannette Bohg and Shuran Song},
  title = {{HoMMI}: Learning Whole-Body Mobile Manipulation from Human Demonstrations},
  journal = {arXiv preprint arXiv:2603.03243},
  year = {2026}
}

@inproceedings{jauhri2024actpermoma,
  author = {Snehal Jauhri and Sophie Lueth and Georgia Chalvatzaki},
  title = {Active-Perceptive Motion Generation for Mobile Manipulation},
  booktitle = {IEEE International Conference on Robotics and Automation (ICRA)},
  year = {2024}
}

@article{huang2024rekep,
  author = {Wenlong Huang and Chen Wang and Yunzhu Li and Ruohan Zhang and Li Fei-Fei},
  title = {{ReKep}: Spatio-Temporal Reasoning of Relational Keypoint Constraints for Robotic Manipulation},
  journal = {arXiv preprint arXiv:2409.01652},
  year = {2024}
}

@inproceedings{wen2025foundationstereo,
  author = {Bowen Wen and Matthew Trepte and Joseph Aribido and Jan Kautz and Orazio Gallo and Stan Birchfield},
  title = {{FoundationStereo}: Zero-Shot Stereo Matching},
  booktitle = {IEEE/CVF Conference on Computer Vision and Pattern Recognition (CVPR)},
  year = {2025}
}

@article{ravi2024sam2,
  author = {Nikhila Ravi and Valentin Gabeur and Yuan-Ting Hu and Ronghang Hu and Chaitanya Ryali and Tengyu Ma and Haitham Khedr and Roman R{\"a}dle and Chloe Rolland and Laura Gustafson and Eric Mintun and Junting Pan and Kalyan Vasudev Alwala and Nicolas Carion and Chao-Yuan Wu and Ross Girshick and Piotr Doll{\'a}r and Christoph Feichtenhofer},
  title = {{SAM 2}: Segment Anything in Images and Videos},
  journal = {arXiv preprint arXiv:2408.00714},
  year = {2024}
}

@article{dharmarajan2025dream2flow,
  author = {Karthik Dharmarajan and Wenlong Huang and Jiajun Wu and Li Fei-Fei and Ruohan Zhang},
  title = {{Dream2Flow}: Bridging Video Generation and Open-World Manipulation with {3D} Object Flow},
  journal = {arXiv preprint arXiv:2512.24766},
  year = {2025}
}

@article{pankert2020perceptivempc,
  author = {Johannes Pankert and Marco Hutter},
  title = {Perceptive Model Predictive Control for Continuous Mobile Manipulation},
  journal = {IEEE Robotics and Automation Letters},
  volume = {5},
  number = {4},
  pages = {6177--6184},
  year = {2020},
  doi = {10.1109/LRA.2020.3010721}
}

@article{liu2024dynamem,
  author = {Peiqi Liu and Zhanqiu Guo and Mohit Warke and Soumith Chintala and Chris Paxton and Nur Muhammad Mahi Shafiullah and Lerrel Pinto},
  title = {{DynaMem}: Online Dynamic Spatio-Semantic Memory for Open World Mobile Manipulation},
  journal = {arXiv preprint arXiv:2411.04999},
  year = {2024}
}

@article{shah2024bumble,
  author = {Rutav Shah and Albert Yu and Yifeng Zhu and Yuke Zhu and Roberto Mart{\'i}n-Mart{\'i}n},
  title = {{BUMBLE}: Unifying Reasoning and Acting with Vision-Language Models for Building-wide Mobile Manipulation},
  journal = {arXiv preprint arXiv:2410.06237},
  year = {2024}
}

@article{guo2026wanda,
  author = {Lingxiao Guo and Huanyu Li and Guanya Shi},
  title = {Worlds in One Demo: A Synthetic Data Engine for Learning Open-World Mobile Manipulation},
  journal = {arXiv preprint arXiv:2607.13154},
  year = {2026}
}

@article{he2026roboreact,
  author = {Shuliang He and Shuai Wang and Bo Yue and Junchi Teng and Changyu Wang and Guiliang Liu},
  title = {{RoboReact}: Agentic Skill Distillation from Generated Egocentric Videos for Generalizable Whole-Body Manipulation},
  journal = {arXiv preprint arXiv:2608.03387},
  year = {2026}
}

@inproceedings{kerr2025rsrd,
  author = {Justin Kerr and Chung Min Kim and Mingxuan Wu and Brent Yi and Qianqian Wang and Ken Goldberg and Angjoo Kanazawa},
  title = {Robot See Robot Do: Imitating Articulated Object Manipulation with Monocular {4D} Reconstruction},
  booktitle = {Proceedings of The 8th Conference on Robot Learning},
  volume = {270},
  series = {Proceedings of Machine Learning Research},
  pages = {587--603},
  year = {2025}
}

@article{bahety2024screwmimic,
  author = {Arpit Bahety and Priyanka Mandikal and Ben Abbatematteo and Roberto Mart{\'i}n-Mart{\'i}n},
  title = {{ScrewMimic}: Bimanual Imitation from Human Videos with Screw Space Projection},
  journal = {arXiv preprint arXiv:2405.03666},
  year = {2024}
}

@article{mandlekar2023mimicgen,
  author = {Ajay Mandlekar and Soroush Nasiriany and Bowen Wen and Iretiayo Akinola and Yashraj Narang and Linxi Fan and Yuke Zhu and Dieter Fox},
  title = {{MimicGen}: A Data Generation System for Scalable Robot Learning using Human Demonstrations},
  journal = {arXiv preprint arXiv:2310.17596},
  year = {2023}
}

@article{xue2025demogen,
  author = {Zhengrong Xue and Shuying Deng and Zhenyang Chen and Yixuan Wang and Zhecheng Yuan and Huazhe Xu},
  title = {{DemoGen}: Synthetic Demonstration Generation for Data-Efficient Visuomotor Policy Learning},
  journal = {arXiv preprint arXiv:2502.16932},
  year = {2025}
}

@article{simeonov2021ndf,
  author = {Anthony Simeonov and Yilun Du and Andrea Tagliasacchi and Joshua B. Tenenbaum and Alberto Rodriguez and Pulkit Agrawal and Vincent Sitzmann},
  title = {Neural Descriptor Fields: {SE(3)}-Equivariant Object Representations for Manipulation},
  journal = {arXiv preprint arXiv:2112.05124},
  year = {2021}
}

@article{simeonov2022rndf,
  author = {Anthony Simeonov and Yilun Du and Lin Yen-Chen and Alberto Rodriguez and Leslie Pack Kaelbling and Tomas Lozano-Perez and Pulkit Agrawal},
  title = {{SE(3)}-Equivariant Relational Rearrangement with Neural Descriptor Fields},
  journal = {arXiv preprint arXiv:2211.09786},
  year = {2022}
}

@article{fu2024mobilealoha,
  author = {Zipeng Fu and Tony Z. Zhao and Chelsea Finn},
  title = {{Mobile ALOHA}: Learning Bimanual Mobile Manipulation with Low-Cost Whole-Body Teleoperation},
  journal = {arXiv preprint arXiv:2401.02117},
  year = {2024}
}

@inproceedings{xu2023xskill,
  author = {Mengda Xu and Zhenjia Xu and Cheng Chi and Manuela Veloso and Shuran Song},
  title = {{XSkill}: Cross Embodiment Skill Discovery},
  booktitle = {Proceedings of The 7th Conference on Robot Learning},
  volume = {229},
  series = {Proceedings of Machine Learning Research},
  pages = {3536--3555},
  year = {2023}
}

@article{ahn2022saycan,
  author = {Michael Ahn and Anthony Brohan and Noah Brown and others},
  title = {Do As I Can, Not As I Say: Grounding Language in Robotic Affordances},
  journal = {arXiv preprint arXiv:2204.01691},
  year = {2022}
}

@article{huang2022innermonologue,
  author = {Wenlong Huang and Fei Xia and Ted Xiao and Harris Chan and Jacky Liang and Pete Florence and Andy Zeng and Jonathan Tompson and Igor Mordatch and Yevgen Chebotar and Pierre Sermanet and Noah Brown and Tomas Jackson and Linda Luu and Sergey Levine and Karol Hausman and Brian Ichter},
  title = {Inner Monologue: Embodied Reasoning through Planning with Language Models},
  journal = {arXiv preprint arXiv:2207.05608},
  year = {2022}
}

@article{liang2022codeaspolicies,
  author = {Jacky Liang and Wenlong Huang and Fei Xia and Peng Xu and Karol Hausman and Brian Ichter and Pete Florence and Andy Zeng},
  title = {Code as Policies: Language Model Programs for Embodied Control},
  journal = {arXiv preprint arXiv:2209.07753},
  year = {2022}
}

@article{huang2023voxposer,
  author = {Wenlong Huang and Chen Wang and Ruohan Zhang and Yunzhu Li and Jiajun Wu and Li Fei-Fei},
  title = {{VoxPoser}: Composable {3D} Value Maps for Robotic Manipulation with Language Models},
  journal = {arXiv preprint arXiv:2307.05973},
  year = {2023}
}

@inproceedings{dong2026jfto,
  author = {Xiaoxiang Dong and Matthew Johnson-Roberson and Weiming Zhi},
  title = {Joint Flow Trajectory Optimization For Feasible Robot Motion Generation from Video Demonstrations},
  booktitle = {IEEE International Conference on Robotics and Automation (ICRA)},
  year = {2026}
}

@inproceedings{barron2026crossinstruct,
  author = {William Barron and Xiaoxiang Dong and Matthew Johnson-Roberson and Weiming Zhi},
  title = {Cross-Modal Instructions for Robot Motion Generation},
  booktitle = {IEEE International Conference on Robotics and Automation (ICRA)},
  year = {2026}
}

@article{zhao2026trimanpolicy,
  author = {James Zhao and Mingyuan Ba and Weiming Zhi},
  title = {Tri-Manual Visuomotor Imitation Learning of Robot Policies},
  journal = {arXiv preprint arXiv:2607.25731},
  year = {2026}
}
\end{document}